# Active Learning for Network Intrusion Detection

Amir Ziai[1]

[1] Stanford University, 450 Serra Mall, Stanford, CA 94305, USA
`amirziai@stanford.com`

**Abstract.** Network operators are generally aware of common attack vectors that they defend against. For most networks the vast majority of traffic is legitimate. However new attack vectors are continually designed and attempted by bad actors which bypass detection and go unnoticed due to low volume. One strategy for finding such activity is to look for anomalous behavior. Investigating anomalous behavior requires significant time and resources. Collecting a large number of labeled examples for training supervised models is both prohibitively expensive and subject to obsoletion as new attacks surface. A purely unsupervised methodology is ideal; however, research has shown that even a very small number of labeled examples can significantly improve the quality of anomaly detection. A methodology that minimizes the number of required labels while maximizing the quality of detection is desirable. False positives in this context result in wasted effort or blockage of legitimate traffic and false negatives translate to undetected attacks. We propose a general active learning framework and experiment with different choices of learners and sampling strategies.

**Keywords:** Network Security, Machine Learning, Active Learning.

# 1    Introduction

Detecting anomalous activity is an active area of research in the security space. Tuor et al. use an online anomaly detection method based on deep learning to detect anomalies. This methodology is compared to traditional anomaly detection algorithms such as Isolation Forests (IF) and a Principal Component Analysis (PCA) based approach and found to be superior. However, no comparison is provided with semi-supervised or active learning approaches which leverage a small amount of labeled data [7]. The authors later propose another unsupervised methodology leveraging Recurrent Neural Network (RNN) to ingest the log-level event data as opposed to aggregated data [8].

Pimentel et al. propose a generalized framework for unsupervised anomaly detection. They argue that purely unsupervised anomaly detection is undecidable without a prior on the distribution of anomalies and learned representations have simpler statistical structure which translate to better generalization. They propose an active learning approach with a logistic regression classifier as the learner [5]. Veeramachaneni et al. propose a human-in-the-loop machine learning system that provides both insights to the analyst as well as addressing large data processing concerns. This system uses unsupervised methods to surface anomalous data points for the analyst to label and a combination of supervised and unsupervised methods to predict the attacks [6]. In this work we also propose an analyst-in-the-loop active learning approach. However, our approach is not opinionated about the sampling strategy or the learner used in active learning. We will explore trade-offs in that design space.

# 2    Dataset

We have used the KDD Cup 1999 dataset which consists of about 500K records representing network connections in a military environment. Each record is either "normal" or one of 22 different types of intrusion such as smurf, IP sweep, and teardrop. Out of these 22 categories only 10 have at least 100 occurrences, the rest were removed. Each record has 41 features including duration, protocol, and bytes exchanged. Prevalence of attack types varies substantially with smurf being the most pervasive at about 50% of total records and Nmap at less than 0.01% of total records:

**Table 1.** Prevalence and number of attacks for each of the 10 attack types

| label | attacks | prevalence | prevalence (overall) | records |
|---|---|---|---|---|
| smurf. | 280790 | 0.742697 | 0.568377 | 378068 |
| neptune. | 107201 | 0.524264 | 0.216997 | 204479 |
| back. | 2203 | 0.022145 | 0.004459 | 99481 |
| satan. | 1589 | 0.016072 | 0.003216 | 98867 |
| ipsweep. | 1247 | 0.012657 | 0.002524 | 98525 |
| portsweep. | 1040 | 0.010578 | 0.002105 | 98318 |
| warezclient. | 1020 | 0.010377 | 0.002065 | 98298 |
| teardrop. | 979 | 0.009964 | 0.001982 | 98257 |
| pod. | 264 | 0.002707 | 0.000534 | 97542 |
| nmap. | 231 | 0.002369 | 0.000468 | 97509 |

## 2.1 Input and output example

The following table depicts 3 rows of data (excluding the label):

**Table 2.** Snippet of input data

| duration | protocol_type | service | flag | src_bytes | dst_bytes | land | wrong_fragment | urgent | hot | ... | dst_host_srv_count |
|---|---|---|---|---|---|---|---|---|---|---|---|
| 0 | tcp | http | SF | 181 | 5450 | 0 | 0 | 0 | 0 | ... | 9 |
| 0 | tcp | http | SF | 239 | 486 | 0 | 0 | 0 | 0 | ... | 19 |
| 0 | tcp | http | SF | 235 | 1337 | 0 | 0 | 0 | 0 | ... | 29 |

The objective of the detection system is to label each row as either "normal" or "anomalous".

## 2.2 Processing pipeline

We generated 10 separate datasets consisting of normal traffic and each of the attack vectors. This way we can study the proposed approach over 10 different attack vectors with varying prevalence and ease of detection. Each dataset is then split into train, development, and test partitions with 80%, 10%, and 10% proportions. All algorithms are trained on the train set and evaluated on the development set. The winning strategy is tested on the test set to generate an unbiased estimate of generalization. Categorical features are one-hot encoded and missing values are filled with zero.

## 3 Approach

### 3.1 Evaluation metric

Since labeled data is very hard to come by in this space, we have decided to treat this problem as an active learning one. Therefore, the machine learning model receives a subset of the labeled data. We will use the F1 score to capture the trade-off between precision and recall:

$$F_1 = \frac{2 \times P \times R}{P + R}$$

where $P = TP/(TP + FP)$, $R = TP/(TP + FN)$, $TP$ is true positives, $FP$ is false positives, and $FN$ is the number of false negatives.

A model that is highly precise (does not produce false positives) is desirable as it won't waste the analyst's time. However, this usually comes at the cost of being overly conservative and not catching anomalous activity that is indeed an intrusion.

### 3.2 Oracle and baseline

Labeling effort is a major factor in this analysis and a dimension along which we'll define the upper and lower bounds of the quality of our detection systems. A purely unsupervised approach would be ideal as there's no labeling involved. We'll use an Isolation Forest [1] to establish our baseline. Isolation Forests (IF) are widely, and very successfully, used for anomaly detection. An IF consists of a number of Isolation Trees, each of which are constructed by selecting random features to split and then selecting a random value to split on (random value in the range of continuous variables or random value for categorical variables). Only a small random subset of the data is used for growing the trees and usually a maximum allowable depth is enforced to curb computational cost. We have used 10 trees for each IF. Intuitively, anomalous data points are easier to isolate with a smaller average number of splits and therefore tend to be closer to the root. The average closeness to the root is proportional to the anomaly score (i.e. the lower this score the more anomalous the data point).

A completely supervised approach would incur maximum cost as we'll have to label every data point. We have used a random forest classifier with 10 estimators trained on the entire training dataset to establish the upper bound (i.e. oracle). The following F1 scores are reported for evaluation on the development set:

**Table 3.** Oracle and baseline for different attack types.

| Label | Baseline F1 | Oracle F1 |
| --- | --- | --- |
| smurf | 0.38 | 1.00 |
| neptune | 0.49 | 1.00 |
| back | 0.09 | 1.00 |
| satan | 0.91 | 1.00 |
| ipsweep | 0.07 | 1.00 |
| portsweep | 0.53 | 1.00 |
| warezclient | 0.01 | 1.00 |
| teardrop | 0.30 | 1.00 |
| pod | 0.00 | 1.00 |
| nmap | 0.51 | 1.00 |
| Mean ± standard deviation | 0.33 ± 0.29 | 1.00 ± 0.01 |

### 3.3 Active learning

The proposed approach starts with training a classifier on a small random subset of the data (i.e. 1,000 samples) and then continually queries a security analyst for the next record to label. There's a maximum budget of 100 queries.

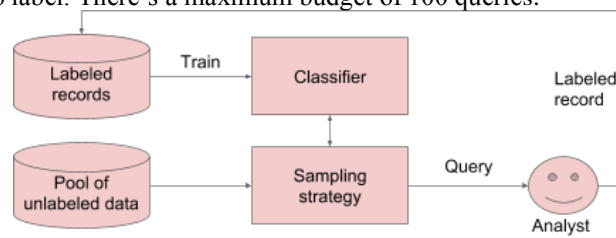

**Fig. 1.** Active learning scheme.

This approach is highly flexible. The choice of classifier can range from logistic regression all the way up to deep networks as well as any ensemble of those models. Moreover, the hyper-parameters for the classifier can be tuned on every round of training to improve the quality of predictions. The sampling strategy can range from simply picking random records to using classifier uncertainty or other elaborate schemes. Once a record is labeled it's removed from the pool of labeled data and placed into the labeled records database. We're assuming that labels are trustworthy which may not necessarily be true. In other words, the analyst might make a mistake in labeling or there may be low consensus among analysts around labeling. In the presence of those issues we'd need to extend this approach to query multiple analysts and to build the consensus of labels into the framework.

## 4 Experiments

### 4.1 Learners and sampling strategies

We used a Logistic Regression (LR) classifier with L2 penalty as well as a Random Forest (RF) classifier with 10 estimators, Gini impurity for splitting criteria, and unlimited depth for our choice of learners. We also chose three sampling strategies. First is a random strategy that randomly selects a data point from the unlabeled pool. The second option is uncertainty sampling that scores the entire database of unlabeled data and then selects the data point with the highest uncertainty. The first option is entropy sampling, which calculates the entropy over the positive and negative classes and selects the highest entropy data point. Ties are broken randomly for both uncertainty and entropy sampling.

The following tabulates the F1 score immediately after the initial training (F1 initial) followed by the F1 score after 10, 50, and 100 queries to the analyst across different learners and sampling strategies aggregated over the 10 attack types:

**Table 4.** Effects of learner and sampling strategy on detection quality and latency.

| Learner | Sampling strategy | F1 initial | F1 after 10 | F1 after 50 | F1 after 100 | Train time (s) | Query time (s) |
|---|---|---|---|---|---|---|---|
| LR | Random |  | 0.76 ±0.32 | 0.79 ±0.31 | 0.86 ±0.17 |  | 0.09 ±0.08 |
| LR | Uncertainty | 0.76± 0.32 | 0.83 ±0.26 | 0.85 ±0.31 | 0.88 ±0.20 | **0.05 ±0.01** | 0.10 ±0.08 |
| LR | Entropy |  | 0.83 ±0.26 | 0.85 ±0.31 | 0.88 ±0.20 |  | **0.08 ±0.08** |
| RF | Random |  | 0.91 ±0.12 | 0.84 ±0.31 | 0.95 ±0.07 |  | 0.09 ±0.07 |
| RF | Uncertainty | 0.90± 0.14 | 0.98 ±0.03 | **0.99 ±0.03** | **0.99 ±0.03** | 0.11 ±0.00 | 0.16 ±0.06 |
| RF | Entropy |  | **0.98 ±0.04** | 0.98 ±0.03 | **0.99 ±0.03** |  | 0.12 ±0.08 |

Random forests are strictly superior to logistic regression from a detection perspective regardless of the sampling strategy. It is also clear that uncertainty and entropy sampling are superior to random sampling which suggests that judiciously sampling the unlabeled dataset can have a significant impact on the detection quality, especially in the earlier queries (F1 goes from 0.90 to 0.98 with just 10 queries). It is important to notice that the query time might become a bottleneck. In our examples the unlabeled pool of data is not very large but as this set grows these sampling strategies have to scale accordingly. The good news is that scoring is embarrassingly parallelizable.

The following figure depicts the evolution of detection quality as the system makes queries to the analyst for an attack with high prevalence (i.e. the majority of traffic is an attack):

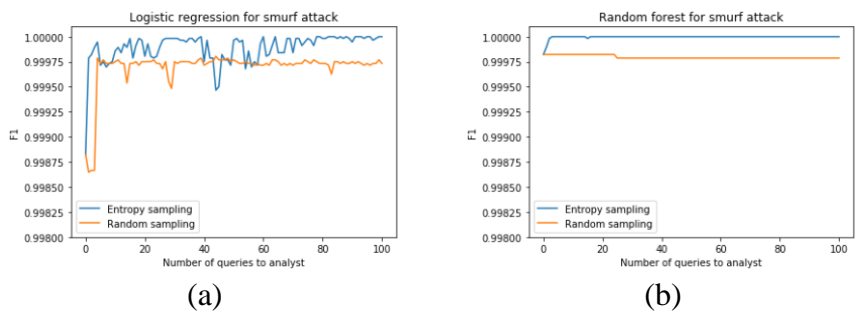

**Fig 2.** Detection quality for a high prevalence attack.

The random forest learner combined with an entropy sampler can get to perfect detection within 5 queries which suggests high data efficiency [2]. We will compare this to the Nmap attack with significantly lower prevalence (i.e. less than 0.01% of the dataset is an attack):

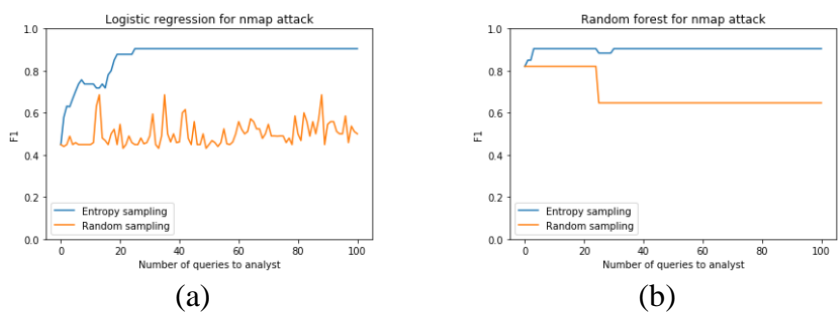

**Fig 3.** Detection quality for a low prevalence attack.

We know from our oracle evaluations that a random forest model can achieve perfect detection for this attack type, however we see that an entropy sampler is not guaranteed to query the optimal sequence of data points. The fact that the prevalence of attacks is very low means that the initial training dataset probably doesn't have a representative set of positive labels that can be exploited by the model to generalize.

The failure of uncertainty sampling has been documented [3] and more elaborate schemes can be designed to exploit other information about the unlabeled dataset that the sampling strategy is ignoring. To gain some intuition into these deficiencies we'll unpack a step of entropy sampling for the Nmap attack. Figure 4 compares (a) the relative feature importance after the initial training to (b) the oracle:

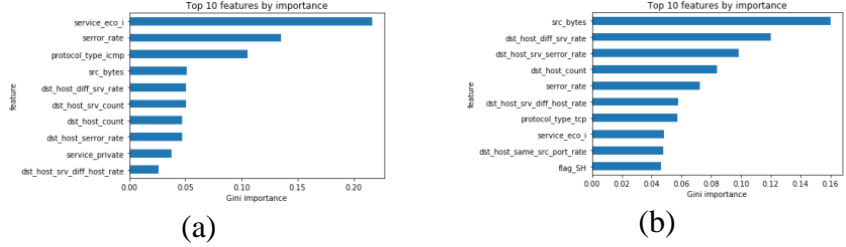

(a)        (b)

**Fig 4.** Random forest feature importance for (a) initial training and (b) oracle.

The oracle graph suggests the "src_bytes" is a feature that the model is highly reliant upon for prediction. However, our initial training is not reflecting this. We'll compute the z-score for each of the positive labels in our development set:

$$z_{f_i} = \frac{\left|\mu_{R_{f_i}} - \mu_{W_{f_i}}\right|}{\sigma_{R_{f_i}}}$$

where $\mu_{R_{f_i}}$ is the average value of the true positives for feature $i$ (i.e. $f_i$), $\mu_{W_{f_i}}$ is the average value of the false positives or false negatives, and $\sigma_{R_{f_i}}$ is the standard deviation of the values in the case of true positives.

The higher this value is for a feature the more our learner need to know about it to correct the discrepancy. However, we see that the next query made by the strategy does not involve a decision around this fact. The score for "src_bytes" is an order of magnitude larger than other features. The model continues to make uncertainty queries staying oblivious to information about specific features that it needs to correct for.

### 4.2 Ensemble learning

Creating an ensemble of classifiers is usually a very effective way to combine the power of multiple learners [4]. This strategy is highly effective when the errors made by classifiers in the ensemble tend to cancel out and are not compounded. To explore this idea, we designed a weighted ensemble:

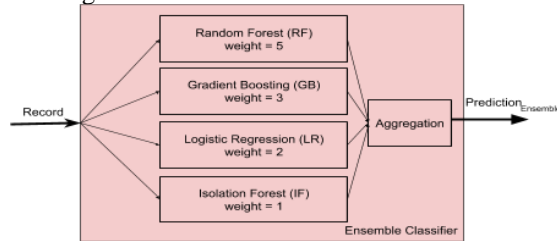

**Fig 5.** Ensemble learner.

The prediction in the above diagram is calculated as follows:

$$Prediction_{Ensemble} = \mathbb{I}[\sum_{e \epsilon E} w_e \, Prediction_E > \sum_{e \epsilon E} w_e / 2]$$

where $Prediction_E \, \epsilon \, \{0,1\}$ is the binary prediction associated with the classifier $e \, \epsilon \, E = \{RF, GB, LR, IF\}$ and $w_e$ is the weight of the classifier in the ensemble.

The weights are proportional to the level of confidence we have in each of the learners. We've added a gradient boosting classifier with 10 estimators.

Unfortunately, the results of this experiment suggest that this particular ensemble is not adding any additional value. Figure 6 shows that at best the results match that of random forest (a) and in the worst case they can be significantly worse (b):

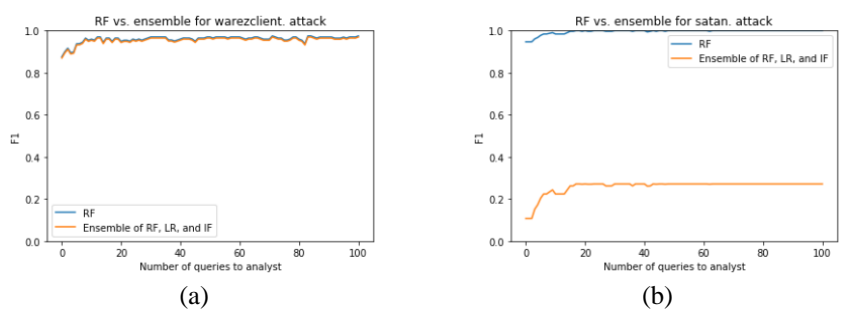

(a) (b)

**Fig 6.** Ensemble active learning results for warezclient and satan attacks.

The majority of the error associated with this ensemble approach relative to only using random forests can be attributed to a high false negative rate. The other four algorithms are in most cases conspiring to generate a negative class prediction which overrides the positive prediction of the random forest.

### 4.3    Sampling the outliers generated using unsupervised learning

Finally, we explore whether we can use an unsupervised method for finding the most anomalous data points to query. If this methodology is successful, the sampling strategy is decoupled from active learning and we can simply precompute and cache the most anomalous data points for the analyst to label.

We compared a sampling strategy based on isolation forest with entropy sampling:

**Table 5.** Active learning with an unsupervised sampling strategy.

| Sampling strategy | Initial F1 | F1 after 10 | F1 after 50 | F1 after 100 |
|---|---|---|---|---|
| Isolation Forest | 0.94 ± 0.07 | 0.94±0.05 | 0.95±0.09 | 0.93±0.09 |
| Entropy | | **0.98±0.03** | **0.99±0.03** | **0.99±0.03** |

In both cases we're using a random forest learner. The results suggest that entropy sampling is superior since it's sampling the most uncertain data points in the context of the current learner and not a global notion of anomaly which isolation forest provides.

## 5     Conclusion

We have proposed a general active learning framework for network intrusion detection. We experimented with different learners and observed that more complex learners can achieve higher detection quality with significantly less labeling effort for most attack types. We did not explore other complex models such as deep neural networks and did not attempt to tune the hyper-parameters of our model. Since the bottleneck associated with this task is the labeling effort, we can add model tuning while staying within the acceptable latency requirements.

We then explored a few sampling strategies and discovered that uncertainty and entropy sampling can have a significant benefit over unsupervised or random sampling. However, we also realized that these strategies are not optimal, and we can extend them to incorporate available information about the distribution of the features for mispredicted data points. We attempted a semi-supervised approach called Label Spreading that builds the affinity matrix over the normalized graph Laplacian which can be used to create pseudo-labels for unlabeled data points [9]. However, this methodology is very memory-intensive, and we couldn't successfully train and evaluate it on all of the attack types.

## References

1. Zhou D, Bousquet O, Lal TN, Weston J, Schölkopf B (2004) Learning with local and global consistency. Advances in neural information processing systems (pp. 321-328)
2. Mussmann S, Liang P (2018) On the relationship between data efficiency and error for uncertainty sampling. arXiv preprint arXiv:1806.06123
3. Zhu J, Wang H, Yao T, Tsou BK (2008) Active learning with sampling by uncertainty and density for word sense disambiguation and text classification. Proceedings of the 22nd International Conference on Computational Linguistics-Volume 1: 1137-1144
4. Zainal A, Maarof MA, Shamsuddin SM (2009) Ensemble classifiers for network intrusion detection system. Journal of Information Assurance and Security 4(3): 217-225
5. Pimentel T, Monteiro M, Viana J, Veloso A, Ziviani N (2018) A generalized active learning approach for unsupervised anomaly detection. arXiv preprint arXiv:1805.09411
6. Veeramachaneni K, Arnaldo I, Korrapati V, Bassias C, Li K (2016) AI^ 2: training a big data machine to defend. Big Data Security on Cloud (BigDataSecurity), IEEE International Conference on High Performance and Smart Computing (HPSC), and IEEE International Conference on Intelligent Data and Security (IDS), IEEE 2nd International Conference: 49-54
7. Tuor A, Kaplan S, Hutchinson B, Nichols N, Robinson S (2017) Deep learning for unsupervised insider threat detection in structured cybersecurity data streams. arXiv preprint arXiv:1710.00811
8. Tuor A, Baerwolf R, Knowles N, Hutchinson B, Nichols N, Jasper R (2018) Recurrent neural network language models for open vocabulary event-level cyber anomaly detection. Workshops at the Thirty-Second AAAI Conference on Artificial Intelligence

9. Zhou D, Bousquet O, Lal TN, Weston J, Schölkopf B (2004) Learning with local and global consistency. Advances in neural information processing systems: 321-328